\documentclass[a4paper,conference]{IEEEtran}
\usepackage{cite}
\usepackage{amsmath}
\usepackage{xcolor}
\usepackage{xspace}
\usepackage{epsfig}
\usepackage{graphicx}
\usepackage{amssymb}
\usepackage{times}  %Required
\usepackage{helvet}  %Required
\usepackage{courier}  %Required
\usepackage{url}  %Required
\usepackage{multirow,floatrow, comment, enumerate}
\usepackage{amsmath,amssymb}
\usepackage{color}
\usepackage{algorithmic}
\usepackage[linesnumbered,ruled,vlined]{algorithm2e}

\ifCLASSINFOpdf
\else
\fi
\begin{document}
\title{Dual-MTGAN: Stochastic and Deterministic Motion Transfer for Image-to-Video Synthesis }

\author{
% \IEEEauthorblockN{Michael Shell}
% \IEEEauthorblockA{School of Electrical and\\Computer Engineering\\
% Georgia Institute of Technology\\
% Atlanta, Georgia 30332--0250\\
% Email: http://www.michaelshell.org/contact.html}
% \and
% \IEEEauthorblockN{Fu-En Yang}
% \IEEEauthorblockA{Graduate Institude of Communication Engineering \\ National Taiwan University\\
% Taipei, Taiwan\\
% Email: r07942077@ntu.edu.tw}
% \and
% \IEEEauthorblockN{James Kirk\\ and Montgomery Scott}
% \IEEEauthorblockA{Starfleet Academy\\
% San Francisco, California 96678--2391\\
% Telephone: (800) 555--1212\\
% Fax: (888) 555--1212}
}


\author{\IEEEauthorblockN{Fu-En Yang$^{1,2}$$^{*}$,
Jing-Cheng Chang$^{1}$$^{*}$,
Yuan-Hao Lee$^{1}$, and
Yu-Chiang Frank Wang$^{1,2}$}
\IEEEauthorblockA{$^{1}$Graduate Institute of Communication Engineering, National Taiwan University, Taiwan \\
$^{2}$ASUS Intelligent Cloud Services, Taiwan \\ Email: \{f07942077, b04901138, r07942074, ycwang\}@ntu.edu.tw}}
% \IEEEauthorblockA{\IEEEauthorrefmark{2}Twentieth Century Fox, Springfield, USA\\
% Email: homer@thesimpsons.com}
% \IEEEauthorblockA{\IEEEauthorrefmark{3}Starfleet Academy, San Francisco, California 96678-2391\\
% Telephone: (800) 555--1212, Fax: (888) 555--1212}
% \IEEEauthorblockA{\IEEEauthorrefmark{4}Tyrell Inc., 123 Replicant Street, Los Angeles, California 90210--4321}}


\maketitle



% \fi
\IEEEpeerreviewmaketitle

\makeatletter
\def\blfootnote{\xdef\@thefnmark{}\@footnotetext}
\makeatother
\blfootnote{* indicates equal contribution.}

\begin{abstract}
Generating videos with content and motion variations is a challenging task in computer vision. While the recent development of GAN allows video generation from latent representations, it is not easy to produce videos with particular content of motion patterns of interest. In this paper, we propose Dual Motion Transfer GAN (Dual-MTGAN), which takes image and video data as inputs while learning disentangled content and motion representations. Our Dual-MTGAN is able to perform deterministic motion transfer and stochastic motion generation. Based on a given image, the former preserves the input content and transfers motion patterns observed from another video sequence, and the latter directly produces videos with plausible yet diverse motion patterns based on the input image. The proposed model is trained in an end-to-end manner, without the need to utilize pre-defined motion features like pose or facial landmarks. Our quantitative and qualitative results would confirm the effectiveness and robustness of our model in addressing such conditioned image-to-video tasks.
\end{abstract}

\section{Introduction}
% Task definition - conditional video generation
Producing video sequences with desirable information is among active research topics in computer vision and machine learning.
%In general, a video sequence can be generated by an input image, a video sequence, or a combination of both. Based on different conditioned inputs, the corresponding output videos can be possibly obtained with particular information of interest.
In general, a video sequence can be generated based on different corresponding conditioned inputs, such as an input image, another video sequence, or a combination of both, depending on the information of interest.
In this work, we address conditional video generation given an input image, which either allows transfer of deterministic motion patterns from another video sequence of interest, or exhibits the ability to produce realistic video outputs with diverse motion patterns.
% Thus, we refer to this task as stochastic conditional video synthesis with deep motion transfer.

% unconditional video generation -> conditional video generation
Several existing works~\cite{vondrick2016generating,saito2017temporal,tulyakov2018mocogan} generate video sequences by randomly sampling latent vectors from some prior distributions (e.g., Gaussian prior), which are observed from training videos. To better manipulate video contents, recent works consider a single image~\cite{li2018flow,pan2019video} or consecutive video frames~\cite{mathieu2015deep,visualdynamics18,villegas2017learning,babaeizadeh2017stochastic,denton2018stochastic,lee2018stochastic,jang2018video,p2pvg2019} as inputs to produce videos. Via modeling motion information in a probabilistic manner, diverse motion patterns may be observed in their output videos (e.g., \cite{li2018flow,babaeizadeh2017stochastic,denton2018stochastic}). Despite promising performances, the above methods cannot easily capture and transfer motion information from another video sequence of interest, which would limit the use of their video generation models. 

Aiming at transferring motion patterns across videos, some recent works either extend Generative Adversarial Networks (GANs)~\cite{goodfellow2014generative} for video retargeting in a frame-by-frame manner\cite{bansal2018recycle}, or leverage explicit information (e.g., pose skeleton, 3D facial model, or key-point) from video sequences of interest with input frames to perform video-to-video translation~\cite{wang2018vid2vid,zhang2019one,wang2019few,zakharov2019few,villegas2018neural,chan2018everybody,zhao2018learning,siarohin2018animating,gu2019flnet}.
Nevertheless, the above models are specifically designed to transfer \textit{deterministic} motion patterns from particular video inputs, and thus lack the flexibility in showing motion stochasticity in the output videos.

\begin{figure}[t!]
	\centering
	\includegraphics[width=1\textwidth]{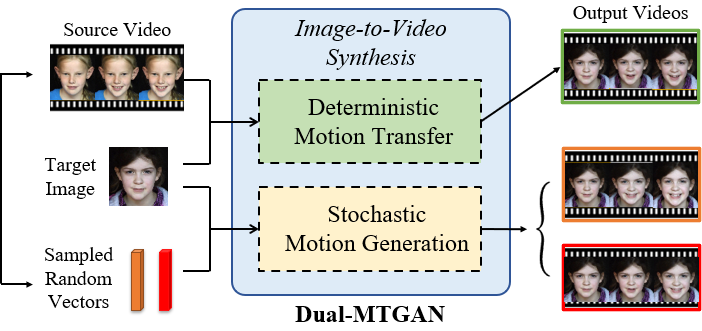}
 	\vspace{2mm}
     \caption{ 
Illustration of our Dual-MTGAN for image-to-video synthesis. Given a target image of interest, the goal is to either produce video outputs with motion diversity, or to generate a video sequence with motion patterns matching those of the source video. Note that motion stochasticity is modeled from source video data during training.
}
	\vspace{-4mm}
	\label{fig:introduction}
\end{figure}

\begin{table*}[!tp]
\centering
\caption{Comparisons of recent works on video synthesis.}
\vspace{1mm}
\resizebox{0.8\textwidth}{!}{
\label{tab:com}
\begin{tabular}{cccccc}
\hline
% \multirow{8}{2}{Market-1501} & \multicolumn{8}{3}{DukeMTMC-reID} \\ 
      & Representation        & Motion        & Without Pre-defined         & Generate Video      & Results with      \\ 
      & Disentanglement    & Transfer  & Prior Information  & from an Image  & Diversity \\\hline

MoCoGAN~\cite{tulyakov2018mocogan}            & ${\surd}$                & -             & ${\surd}$  & (${\surd}$) & -            \\
MCNet~\cite{villegas2017decomposing}            & ${\surd}$                & -             & ${\surd}$  & - & -            \\
DRNet~\cite{denton2017unsupervised}            & ${\surd}$                & ${\surd}$             & ${\surd}$  & - & -            \\ 
SVG~\cite{denton2018stochastic}            & -                & -             & ${\surd}$ & ${\surd}$ & ${\surd}$       \\ 
Chan~\textit{et al.}~\cite{chan2018everybody}             & - & ${\surd}$ & - & - & -                                                      \\
Vid2vid~\cite{wang2018vid2vid}           & -                       & ${\surd}$               & - & - & -            \\
Recycle-GAN~\cite{bansal2018recycle}        & -                       & ${\surd}$               & ${\surd}$ & - & -             \\  
FR-GAN~\cite{zhao2018learning}            & -                    & ${\surd}$             & - & ${\surd}$ & -          \\ 
Li~\textit{et al.}~\cite{li2018flow}            & -                & -             & - & ${\surd}$ & ${\surd}$        \\ 
seg2vid~\cite{pan2019video}            & -                & -             & - & ${\surd}$ & -  \\
X2Face~\cite{wiles2018x2face}  & -                    & ${\surd}$             & - & ${\surd}$ & -          \\ 
Monkey-Net~\cite{siarohin2018animating}             & -                    & ${\surd}$             & - & ${\surd}$ & -          \\ 
\textbf{Ours} (Dual-MTGAN)      & ${\surd}$ & ${\surd}$ & ${\surd}$ & ${\surd}$ & ${\surd}$ \\                 
\hline
\end{tabular}
\vspace{-4mm}
}
\end{table*}

In this paper, we propose a \textit{Dual Motion Transfer GAN} (Dual-MTGAN), which addresses image-to-video synthesis with joint capabilities of stochastic motion generation and deterministic motion transfer (as depicted in Figure~\ref{fig:introduction}). Our Dual-MTGAN disentangles input image and video data into latent representations of content and motion features, which describe visual appearance and model dynamic motion patterns, respectively. While we advance inherent temporal coherence as self-supervision for content feature disentanglement, the motion features are learned via cycle-consistency of motion features during translation. With the decomposed content/motion representations, our model animates the target image with the transferred motion patterns of the source video in a \emph{deterministic} manner by combining the content/motion features extracted from image/video inputs. Moreover, to further exploit the \emph{stochasticity} of motion dynamics in video sequences, the \emph{motion latent space} encoded from source videos would fit a prior distribution. This allows out Dual-MTGAN to synthesize realistic yet diverse output videos by sampling motion features. We note that, our Dual-MTGAN utilizes adversarial learning strategies, which further preserve the plausibility and continuity of the generated video outputs. In the experiments, we show that our model not only transfers motion patterns across videos, but is also able to synthesize realistic videos with motion diversity given a single input image.

%we transfer motion pattern across videos by using a target image of interest associated with the motion pattern extracted from another video, while sample the random vector as the motion information with an input target image to generate reasonable video sequence with reality and stochasticity. To achieve our goal, we perform multi-task learning for considering the both tasks jointly. Our model learns to encode the motion pattern from another video sequence to a latent space and fit a prior distribution (e.g., Gaussian distribution) using a generative model. In this way, we are able to generate multiple consecutive frames from a still image via sampling a random vector, which representing a reasonable motion pattern for whole video sequence, from prior distribution.
%Moreover, we further introduce consistency regularization for motion representation to facilitate to capture the complete motion distribution (i.e., avoid mode collapse), which make the output video with diversity by sampling different random vectors. 
%Finally, we perform an advanced adversarial training strategy both for whole video and individual frame, which ensure the temporal coherent to let generated output closed to the manifold of real video and gain the more plausible result with correct appearance.  

%In the section of related works, we will present a comparison in Table~\ref{tab:com} among the state-of-the-art conditional video generation methods and detailed in section~\ref{sec:related}.  

The contributions of this paper are highlighted below:

\begin{itemize}
\item We address image-to-video generation with flexibility in controlling motion information. Given an input image, our proposed model allows transfer of motion patterns from video data, or synthesis of video sequences with motion diversity. 

\item
By enforcing appearance coherence and motion consistency, our Dual-MTGAN factorizes visual latent representations into disjoint features describing content and motion features in a self-supervised manner.

\item
Our proposed model is simply trained by observing a sourced-domain video and a target-domain image in an end-to-end manner. No auxiliary information or supervision like facial landmarks or shape models.
% Our proposed model is trained in an end-to-end manner, without the need to utilize auxiliary or pre-defined feature information like pose, skeleton, or optical flow.

\end{itemize}

\section{Related Work}\label{sec:related}

\noindent\textbf{Stochastic Video Synthesis.}
Based on GAN~\cite{goodfellow2014generative} architecture, several works~\cite{vondrick2016generating,saito2017temporal,tulyakov2018mocogan} generate video frames from prior distributions observed from training data. For example, {Vondrick~\textit{et al.}~\cite{vondrick2016generating}} and {Saito~\textit{et al.}~\cite{saito2017temporal}} proposed VGAN and TGAN respectively, to learn a mapping between video data and the associated latent spaces.
While they can produce video sequences similar to real ones, they cannot synthesize videos conditioned on the content of interest.    
With the aim of controlling the content of synthesis videos, recent works~\cite{li2018flow,pan2019video} further perform image-to-video generation, which produce plausible video outputs based on a single image. For example, {Li \textit{et al.}~\cite{li2018flow}} proposed a two-stage training framework incorporating generative models with optical flow supervision to generate video frames with diversity. 
% {Pan \textit{et al.}~\cite{pan2019video}} utilized image-to-image translation followed by integrating optical flow and semantic information for video synthesis.
Nevertheless, these works typically require pre-defined motion priors like optical flow, which hampers the motion diversity and cannot be easily extended to motion transfer.

% \noindent\textbf{Video Prediction and Image-to-Video Generation.}
% Video prediction~\cite{mathieu2015deep,visualdynamics18,villegas2017learning,babaeizadeh2017stochastic,denton2018stochastic,lee2018stochastic,jang2018video} aims at predicting future frames conditioned on consecutive input frames. To extend from deterministic to stochastic settings, a number of works~\cite{visualdynamics18,babaeizadeh2017stochastic,denton2018stochastic,lee2018stochastic} apply generative models~\cite{kingma2013auto,goodfellow2014generative} for modeling motion stochasticity for video prediction. With diverse motion information properly modeled, recent works~\cite{li2018flow,pan2019video} further perform image-to-video generation, which produce plausible video outputs based on a single image. For example, {Li \textit{et al.}~\cite{li2018flow}} proposed a two-stage training framework incorporating generative models with optical flow supervision to generate video frames with diversity. {Pan \textit{et al.}~\cite{pan2019video}} utilized image-to-image translation followed by integrating optical flow and semantic information for video synthesis. Nevertheless, these works typically require explicit motion priors like optical flow, and cannot be easily extended to motion transfer across videos.

\noindent\textbf{Deterministic Motion Transfer.}
The goal of motion transfer is to translate a source video into an output one, in which the motion information is derived from the source video while the visual appearance is preserved in the target output~\cite{bansal2018recycle,wang2018vid2vid,villegas2018neural,chan2018everybody,zhao2018learning,wiles2018x2face,siarohin2018animating,fan2019controllable,shen2019facial}. Video-to-video translation is among the solutions, which directly learns mapping across videos. For example, {Bansal~\textit{et al.}~\cite{bansal2018recycle}} extended CycleGAN~\cite{zhu2017unpaired} and combined spatiotemporal constraints to translate one video into another in a frame-by-frame manner. Vid2vid~\cite{wang2018vid2vid} aims to translate a sequence of semantic representation to consecutive video frames. While satisfactory results have been reported, the above methods are only applicable to translation between two domains (e.g., John	Oliver to Stephen Colbert). Thus, they have limited flexibility in practical scenarios. Another group of works learned to use explicit information as prior information like skeleton or keypoints~\cite{villegas2018neural,chan2018everybody,zhao2018learning,wiles2018x2face,siarohin2018animating} to address this task. For example, Monkey-Net~\cite{siarohin2018animating} incorporated keypoints information to realize motion transfer and also applied the extracted keypoints to generate videos from an image. However, the above assumptions with deterministic settings cannot perform video generation with sufficient motion diversity.

\begin{figure*}[t!]
	\centering
	\includegraphics[width=0.9\textwidth]{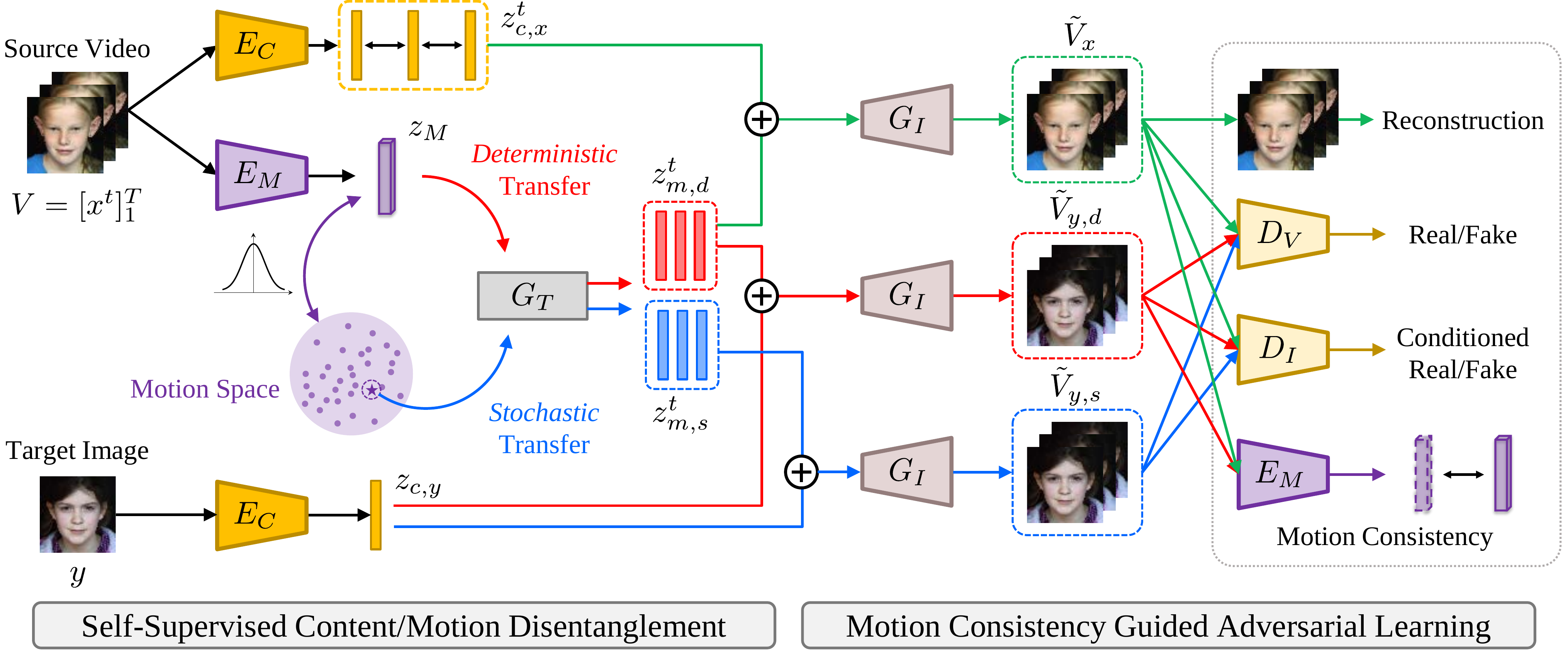}
 	\vspace{2mm}
    \caption{The network architecture of our proposed Dual-MTGAN, which consists of a content encoder $E_C$, a motion encoder $E_M$, a RNN-based motion generator $G_T$, and an image generator $G_I$, with video and image-based discriminators $D_V$ and $D_I$. Note that $z_c$, $z_M$, and $z_m^t$ denote time-invariant content, video motion, and frame-based motion features, respectively. Note that $\oplus$ indicates concatenation of content and motion features. Best viewed in color.}
    \vspace{-2mm}
	\label{fig:archi}
\end{figure*}

\noindent\textbf{Representation Disentanglement for Video Synthesis.}
Aiming at learning interpretable data representations~\cite{chen2016infogan, acgan, lee2018diverse, huang2018multimodal, DRIT_plus, yang2019multi}, the idea of representation disentanglement has been applied to video generation~\cite{villegas2017decomposing,denton2017unsupervised,hsieh2018learning}. For example, Tulyakov~\textit{et al.}~\cite{tulyakov2018mocogan} presented MoCoGAN to generate videos from decomposed random noise representing motion and content information respectively. Note that, while MoCoGAN present an extension version for image-to-video synthesis, they fail to capture and transfer the motion patterns across videos.
% {Villegas \textit{et al.}~\cite{villegas2017decomposing}} proposed MCNet to extract motion across video frames for video prediction.
% DDPAE~\cite{hsieh2018learning} applied structured probabilistic models to decompose video feature for prediction purposes.
Moreover, DRNet~\cite{denton2017unsupervised} disentangled the latent representation into content and pose, and performed motion transfer across videos.
Despite promising prediction and motion transfer results, the above methods lack the ability to exhibit motion stochasticity during the transfer process, and cannot be directly applied to transfer motion across videos.

\section{Proposed Method}\label{sec:method}

\subsection{Problem Definition and Model Overview}\label{ssec:definition}
With the goal of generating videos with plausible content and motion information, we propose a novel Dual-Motion Transfer GAN (Dual-MTGAN) to produce video sequences via observing a combination of source video $V={[x^t]}_1^T$ with $T$ input frames and a target image $y$, or only a given target image $y$. With the content observed from either input domain, our output video would directly transfer motion information from $V$ or exhibit sufficiently realistic motion patterns. 

As illustrated in Figure~\ref{fig:archi}, our Dual-MTGAN utilizes $E_C$ and $E_M$ as content and motion encoders, producing time-invariant content features $z_{c}$ and video dynamic motion features $z_M$, respectively. For dual motion transfer purposes, we generate frame-based motion features $z_{m,d}^t$ or $z_{m,s}^t$ by either directly decoding $z_M$ via a RNN-based motion generator $G_T$, or utilizing feature sampled from the motion latent space as input to $G_T$.
Moreover, the decomposed content and motion representations are learned with both data recovery and plausibility objectives, which are realized by an image generator $G_I$, a video-based discriminator $D_V$, and an image-based discriminator $D_I$. With such adversarial learning strategies, we are able to produce a video sequence based on the target image $y$, with motion patterns either encoded from the source video $V$ or sampled from prior distributions. 
% The details of our proposed network will be discussed in the following subsections.          

\subsection{Self-Supervised Disentanglement of Content/Motion Representations}\label{ssec:disen}
As illustrated in Figure~\ref{fig:archi}, we have a content encoder $E_C$ to extract content features from either source video or target image data.
%\textcolor{blue}{Considering that the appearance has temporally coherent nature across all frames of a video. We take this property as a \emph{self-supervision} to extract content features of source video to be time-invariant (i.e., $z_{c, x}^t=z_{c, x}$).}
As the appearance in a video sequence is temporally coherent across all frames, we leverage this property as a \emph{self-supervision} to ensure that content features extracted from the source video are time-invariant (i.e., $z_{c, x}^t=z_{c, x}$).
In other words, we enforce temporal consistency between cross-frame content features $z_{c, x}^t$ and $z_{c, x}^{t+1}$, which correspond to the loss function $\mathcal{L}_{C}$ defined as follows:
\begin{equation}\label{eq:con}
\begin{aligned}
\mathcal{L}_{C} = {||E_C(x^{t}) - E_C({x}^{t+1})}||_{1}.\\
\end{aligned}
\end{equation}
By enforcing the above loss, our $E_C$ extracts time-invariant content features from input frames, and thus can be applied to extract content features from (target) images as well.

In order to encode motion information observed from video data, we utilize a separate motion encoder $E_M$ with spatial-temporal convolutional architectures (i.e., 3D convolution)~\cite{tran2015learning} in our proposed framework. Moreover, to better exploit the motion features
and allow stochastic sampling during inference, we train our network module in a generative manner, and establish a motion latent space to model the distribution of inherent motion dynamics. More precisely, this is realized by enforcing the \textit{Kullback-Leibler} divergence to encourage the distribution of extracted video motion feature $z_M$ to fit a prior Gaussian distribution $\mathcal{N}(\textbf{0},\textbf{I})$. Thus, the corresponding objective function $\mathcal{L}_{KL}$ is defined as:
\begin{equation}\label{eq:kl}
\begin{aligned}
\mathcal{L}_{KL} = \mathbb{E}[KL(\mathcal{P}(z_M)||\mathcal{N}(\textbf{0},\textbf{I}))]. \\
\end{aligned}
\end{equation}

\noindent where $\mathcal{P}(z_M)$ denotes the distribution of $z_M$. Note that, with the learned motion latent space, we are able to perform dual deterministic and stochastic motion transfer by using motion representations that are either directly encoded or stochastically sampled. 

To generate frame-based motion features for each output frame, we then apply recurrent neural network modules of \textit{LSTM}~\cite{hochreiter1997long} as our frame-based motion generator $G_T$, which learns the distribution of motion latent feature trajectories and outputs $z_{m,d}^t$ or $z_{m,s}^t$ across frames from encoded or sampled motion as input of $G_T$, respectively.

By deploying image generator $G_I$ to synthesize the output frames via concatenating $z_{m,d}^t$ or $z_{m,s}^t$, we perform image-to-video generation conditioned on either content vector $z_{c, x}$ (from source video) or $z_{c, y}$ (from target image). We denote the recovered video sequence from $x$ as $\tilde{V}_x$, and the generated videos from $y$ with encoded or sampled motion as $\tilde{V}_{y,d}$ and $\tilde{V}_{y,s}$, respectively. To be more specific, the output videos are expressed as: 
\begin{equation}\label{eq:v_x}
\begin{aligned}
\tilde{V}_x&=G_I([z_{c,x}^{t}]_1^T, [z_{m,d}^t]_1^T), \\ \tilde{V}_{y,d}&=G_I(z_{c,y}, [z_{m,d}^t]_1^T), \ \tilde{V}_{y,s}=G_I(z_{c,y}, [z_{m,s}^t]_1^T),
\end{aligned}
\end{equation}

% \begin{equation}\label{eq:v_x}
% \begin{aligned}
% \tilde{V}_x=G_I([z_{c,x}^{1:T}, z_m^{1:T}])=\begin{bmatrix}
% \begin{bmatrix}
% z_{c,x}^1\\z_m^1 

% \end{bmatrix}, & \ldots, & \begin{bmatrix}
% z_{c,x}^T\\z_m^T

% \end{bmatrix}
% \end{bmatrix}, \\
% \end{aligned}
% \end{equation}
% \begin{equation}\label{eq:v_y}
% \begin{aligned}
% \tilde{V}_y=G_I([z_{c,y}, z_m^{1:T}])=\begin{bmatrix}
% \begin{bmatrix}
% z_{c,y}\\z_m^1 

% \end{bmatrix}, & \ldots, & \begin{bmatrix}
% z_{c,y}\\z_m^T

% \end{bmatrix}
% \end{bmatrix}. \\
% \end{aligned}
% \end{equation}

The plausibility of the generated output $\tilde{V}_{y,d}$ and $\tilde{V}_{y,s}$~\eqref{eq:v_x} is enforced via adversarial training (detailed in the next subsection), while the recovered output $\tilde{V}_x$ in~\eqref{eq:v_x} can be constrained by the reconstruction loss $\mathcal{L}_{rec}$ defined as:
%While we enforce the plausibility of the generated output $\tilde{V}_y$~\eqref{eq:v_y} via adversarial learning (detailed in the next subsection), we observe the recovered output $\tilde{V}_x$ in~\eqref{eq:v_x} by calculating the reconstruction loss $\mathcal{L}_{rec}$:
\begin{equation}\label{eq:rec}
\begin{aligned}
\mathcal{L}_{rec} = {||\tilde{V}_x - V}||_{1}. 
\end{aligned}
\end{equation}

Through $\mathcal{L}_{rec}$, we not only guarantee $G_I$ with the ability of reconstruction, but also enforce the derivation of disjoint time-variant motion information from the input video to achieve disentanglement of latent representations. 

% Note that, while the above video recovery or generation process utilizes the extracted time-variant dynamic information via $z_m^t$, one can also perform random sampling of $z_M$ for synthesizing diverse yet plausible video outputs. 

%As depicted in Figure~\ref{fig:archi}, this design allows our model to output motion information by either decoding $z_M$ observed by a source-domain video sequence, or by randomly drawing a sample of $z_M$ from normal distribution. Together with the conditioned content feature $z_c$ observed from either the source-domain video or target-domain input image, we will be able to produce desirable video outputs. 

% During training process, the video motion latent representation $z_M$ can be either extracted from source video or sampled from prior distribution

\subsection{Motion-Consistency Guided Adversarial Learning}\label{ssec:trans}
With our derived time-invariant content feature $z_c$ and time-variant motion feature $z_{M}$, we advance adversarial learning strategies to ensure realistic video outputs with appearance guarantees. Moreover, cycle-consistency for motion representations is enforced to facilitate the preservation of learned motion information. We now describe the details below. \\

% As shown in Figure~\ref{fig:archi}, we deploy an image-based discriminator $D_I$, and video-based discriminator $D_V$. To fully exploit the encoded motion features for motion transfer in image-to-video generation, ....
% With the content feature extracted either from source video or target image, the output video is reconstructed at or generated into the appearance of interest. 

\textbf{Appearance-Aware Visual Realism.} We note that, when generating output videos with the appearance of interest, there is no guarantee that the output video would adequately satisfy the appearance information based on the given target image. Hence, as shown in Figure~\ref{fig:archi}, we deploy in our proposed network an image-based discriminator $D_I$ which takes a pair of images as input, and a video-based discriminator $D_V$ which observes an entire video. Inspired by~\cite{reed2016generative}, we design our $D_I$ as a conditional discriminator by introducing an adversarial loss with appearance-aware terms to further confirm that content information can be properly encoded/recovered. This loss function encourages image-based discriminator $D_I$ to not only distinguish the generated video outputs from real ones, but also let $D_I$ identify the mismatch between the generated and the conditioned inputs. In this way, we guarantee the plausibility of generated video frames, while ensuring the content of the output to match the conditioned image (i.e., the first frame of the source video ($x^1$), or the target image ($y$)). For the sake of simplicity, we use $\tilde{V}_y$ to denote either one of $\tilde{V}_{y,d}$ or $\tilde{V}_{y,s}$.  
The objective functions for the above learning process can be defined as follows:
\begin{equation}
\mathcal{L}_{GAN}^I = \mathcal{L}_{GAN,x}^{I} + \mathcal{L}_{GAN,y}^{I}, \,\text{where}     
\end{equation}
\begin{equation*}
\begin{aligned}
\mathcal{L}_{GAN,x}^{I}  = \log(D_{I}(x^1, S_I(V))) 
&+ \frac{1}{2}[\log(1 - D_{I}(x^1, S_I(\tilde{V}_x)))  \\
&+ \log(1 - D_{I}(y, S_I({V})))]\\
\end{aligned}
\end{equation*}
\begin{equation*}
\begin{aligned}
\mathcal{L}_{GAN,y}^{I} = \log(D_{I}(y, y))
&+ \frac{1}{2}[\log(1 - D_{I}(y, S_I(\tilde{V}_y))) \\
&+ \log(1 - D_{I}(x^1, S_I({V}_y)))],\\
\end{aligned}
\end{equation*}
where $S_I$ is an access function which randomly samples a frame from the output video, and $V_y$ is the real video with the same appearance as $y$. We note that, in both $\mathcal{L}^{I}_{GAN,x}$ and $\mathcal{L}^{I}_{GAN,y}$, the first term indicates the ``real'' pairs. The second term represents the ``fake'' pairs, where the first subterm ensures the plausibility and the second subterm avoids content mismatch.

As for video-level adversarial learning, we apply a video-based discriminator $D_V$ to ensure both visual quality and temporal continuity of the entire video output. Thus, the objective function of $D_V$ is calculated as:
\begin{equation}
\mathcal{L}_{GAN}^V = \mathcal{L}_{GAN,x}^V + \mathcal{L}_{GAN,y}^V, \,\text{where}     
\end{equation}
\begin{equation*}\label{eq:vgan}
\begin{aligned}
\mathcal{L}_{GAN,x}^V
= \log(D_{V}(V)) + \log(1 - D_{V}(\tilde{V}_x)),\\ 
\mathcal{L}_{GAN,y}^V 
= \log(D_{V}(V_y)) + \log(1 - D_{V}(\tilde{V}_y)).\\
\end{aligned}
\end{equation*} 
\\
\textbf{Cycle-Consistency for Motion Features.} While the quality and continuity of produced videos are enforced by the above deployment of image and video-based discriminators, they do not explicitly guarantee the disentanglement or modeling of motion information from input videos. To further enrich the derivation of motion space, we additionally advance a motion consistency constraint for preserving motion dynamics between the encoded motion vector $z_M$ and the reconstructed video $\tilde{V}_x$ (or the generated one $\Tilde{V}_{y,d}$). 
We calculate such consistency losses at feature level, which suggests the associated objective function $\mathcal{L}_{M}$ as:
\begin{equation}\label{eq:motion}
\begin{aligned}
\mathcal{L}_{M}  = {||E_M(\tilde{V}_x) - z_M}||_{1} 
+ {||E_M(\Tilde{V}_{y,d}) - z_M}||_{1}.\\
\end{aligned}
\end{equation}

\begin{figure*}[t!]
	\centering
	\includegraphics[width=0.95\textwidth]{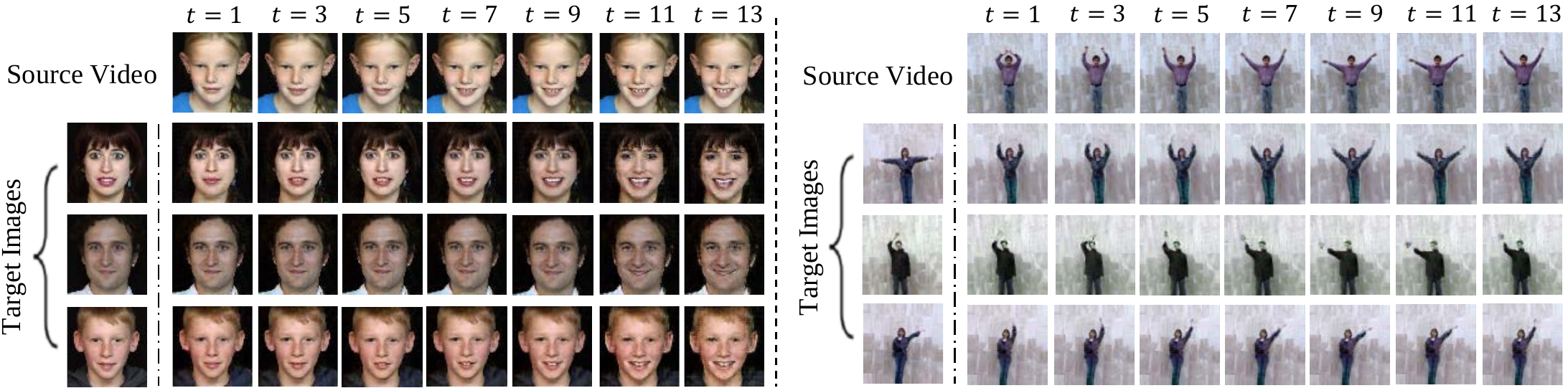}
	\vspace{2mm}
     \caption{Example results of deterministic motion transfer using \textit{facial expression} (left) and \textit{human actions} (right) video data, showing that our model produces videos with content based on the target image while the motion patterns matching those in the source video.}
	\vspace{-4mm}
	\label{fig:exp_mt}
\end{figure*}

\begin{figure*}[t!]
	\centering
	\includegraphics[width=0.95\textwidth]{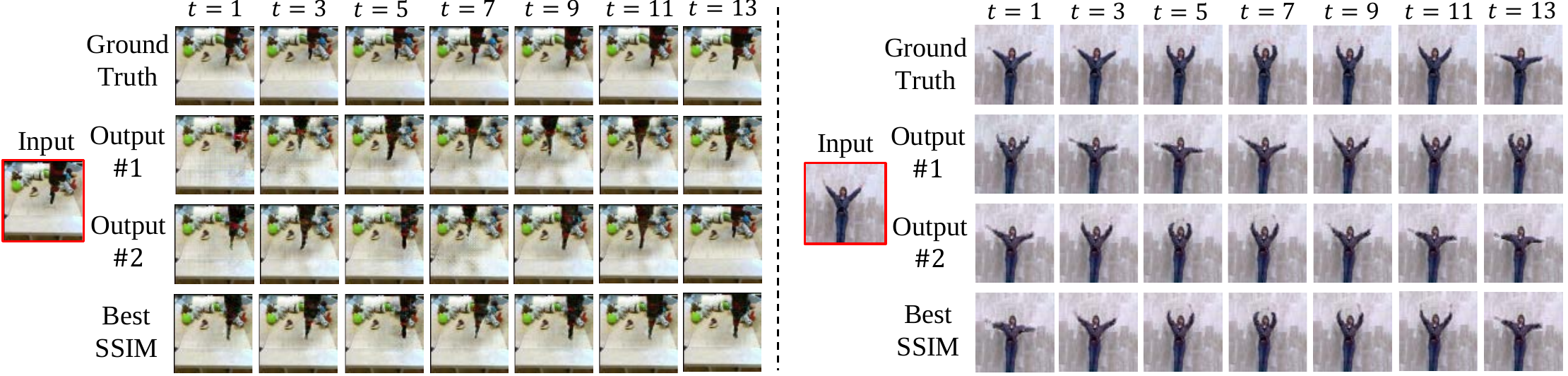}
	\vspace{2mm}
     \caption{Example results of stochastic motion generation using \textit{robot pushing} (left) and \textit{human actions} (right) video data. The frame bounded in red denotes the input frame, and each row shows ground truth or video outputs with motion stochasticity. The last row shows the video with the highest SSIM matching the ground truth video.}
	\vspace{-4mm}
	\label{fig:exp_i2v}
\end{figure*}

\subsection{Full Objectives}\label{ssec:full}
The full objective function of learning our Dual-MTGAN can be summarized as below:
\begin{equation}\label{eq:full}
\begin{aligned}
\displaystyle \min_{E_C, E_M, G_T, G_I} \displaystyle \max_{D_I, D_V} &\mathcal{L} =  \lambda_{1}(\mathcal{L}_{C} + \mathcal{L}_{rec} + \mathcal{L}_{M}) \\
&+ \lambda_{KL}\mathcal{L}_{KL} + \lambda_{I}\mathcal{L}_{GAN}^I + \lambda_{V}\mathcal{L}_{GAN}^V,\\
\end{aligned}
\end{equation}
\noindent In all our experiments, we set the hyperparameters as follows: $\lambda_{1}=10$, $\lambda_{KL}=10^{-2}$, $\lambda_{V}=1$, and $\lambda_{I}=10^{-4}$. Please refer to the supplementary materials for implementation details. 

% \begin{figure}[t!]
% 	\centering
% 	\includegraphics[width=0.96\columnwidth]{Figures/mt_com_v0.png}
% 	\vspace{-2mm}
%      \caption{Qualitative comparisons of deterministic motion transfer with Monkey-Net using \textit{human actions} dataset. }
% 	\vspace{-4mm}
% 	\label{fig:mt_com}
% \end{figure}

%We note that our model is trained in an end-to-end manner without extracting pre-defined or explicit features like pose, skeleton, or optical flow.
Once the learning process is complete, our Dual-MTGAN can be applied to deterministic motion transfer and stochastic motion generation:
\\
1) Given a source video $V$ and a target image $y$, we utilize $E_M$ and $E_C$ to capture the motion feature $z_M$ from $V$ and visual content $z_{c,y}$ from $y$, respectively. We then derive frame-based motion features $z_{m,d}^t$ through $G_T$.
After concatenating with $z_{c,y}$, the image generator $G_I$ is applied to output a video with visual appearance of $y$ and motion patterns of $V$. \\
2) With the input of a target image $y$, we extract its content feature $z_{c}$ by $E_C$, with a sampled motion feature $z_M$ for producing frame-based motion trajectories from $G_T$. By combining $z_{m,s}^t$ with $z_c$, the image generator $G_I$ would output videos with motion diversity.

\section{Experiments}\label{sec:result}

\subsection{Datasets}\label{ssec:dataset}
%We consider three different categories of video datasets, i.e., facial expression, human actions, and robot pushing for performance evaluation:\\
\textbf{Facial expression.} The UvA-Nemo dataset~\cite{dibekliouglu2012you} contains 1240 videos of several identities. Following~\cite{siarohin2018animating}, each frame is resized to $64\times64\times3$ pixels and each video is divided into 16-frame length sequences. Also, we use 1110 videos for training and 124 for inference.

\noindent\textbf{Human actions.} The Weizmann Human Action dataset~\cite{blank2005actions} consists of 90 videos
covering 10 action categories performed by 9 people. We cropped each frame to center on the person and resize the frames to the size of $64\times64\times3$. Following~\cite{tulyakov2018mocogan}, we split the first $\frac{2}{3}$ frames of each video for training, and the remaining $\frac{1}{3}$ frames of video for inference.

% Then we sample 20 mini-clips of length-10 from each training video to form a final training set.

\noindent\textbf{Robot pushing.} The BAIR robot pushing dataset~\cite{ebert2017self} is composed of videos presented by a robotic arm pushing various objects over a table. It contains 40960 videos for training and 256 videos for testing. Each frame is $64\times64\times3$ pixels and each video has 30 frames. 
% Note that, the movement of robot arm is highly stochastic, hence, this dataset can be viewed as a good evaluation for stochastic image-to-video generation.

\begin{figure*}[t!]
	\centering
	\includegraphics[width=0.95\textwidth]{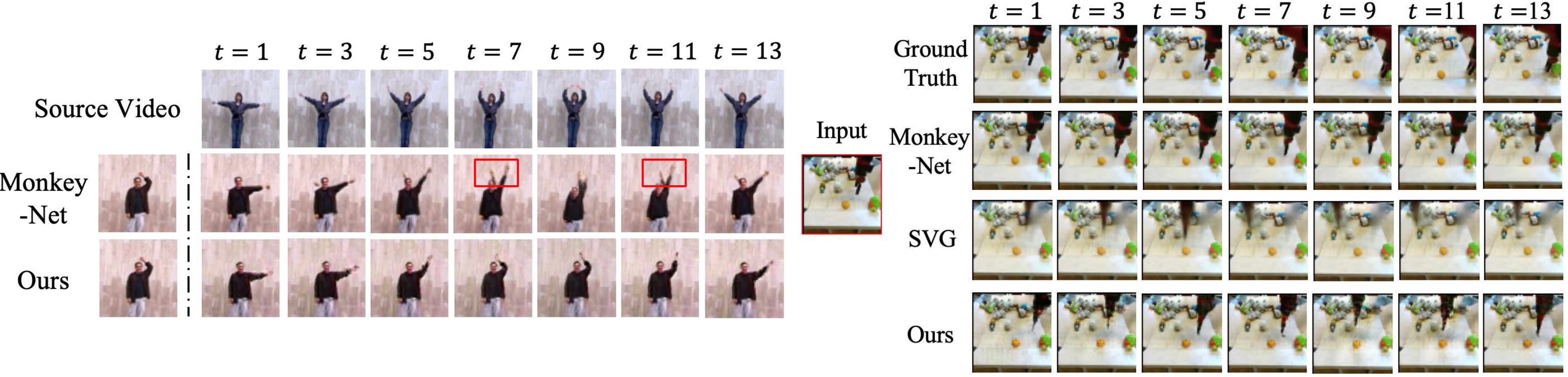}
	\vspace{2mm}
     \caption{Qualitative comparisons with SVG and Monkey-Net. Left part: comparisons of deterministic motion transfer; Right part: comparisons of stochastic motion generation.   }
	\vspace{-2mm}
	\label{fig:com}
\end{figure*}

\subsection{Deterministic Motion Transfer}\label{ssec:exp_retarget}
\subsubsection{Qualitative results and comparisons.} We demonstrate the ability of our Dual-MTGAN in realizing deterministic motion transfer across videos using \textit{facial expression} and \textit{human actions} data. 
% Since no ground truth information is available for the motion transfer outputs, we show example results in Figure~\ref{fig:exp_mt}. 
As described in Figure~\ref{fig:archi}, given a target image with a source video, our model is able to synthesize a corresponding video with motion patterns transferred from those observed in the source video. In Figure~\ref{fig:exp_mt}, we see that we were able to generate output videos whose appearances were properly preserved from the input target image, while the motion patterns were consistent to those of the source videos. Take the right part of Figure~\ref{fig:exp_mt} for example, despite the poses of target images were different from those in the source videos (e.g., the person in the source video is waving both of his hands, while a different person in the target image is raising one hand only), our model was able to preserve the content of the target image and successfully extracted/transferred the motion pattern across videos (e.g., the hand(s) waving up or down).

In the \emph{left} part of Figure~\ref{fig:com}, we compare our model with Monkey-Net~\cite{siarohin2018animating} (a state-of-the-art deterministic motion transfer model) to show superior visual realism achieved by our Dual-MTGAN. Since Monkey-Net integrates keypoints information to perform motion transfer, its output may exhibit inferior visual realism when the appearance of the target image is largely different from that of the source video. A number of frames in the output of Monkey-Net contain irrational limbs resulted from keypoints extracted from the source video (highlighted in red blocks), leading to unrealistic and blurry outputs.
From the above experiments, we see that our Dual-MTGAN exhibits capabilities in performing image-to-video generation with deterministic motion transfer.

% We observed that our model performs more realistic and reasonable motion pattern. We note that, due to Monkey-Net integrates keypoints information to perform motion transfer, it typically suffers from transferring the motion patterns extracted from the appearance with large variance.

\begin{table}[!tp]
\centering
\caption{Quantitative comparisons of stochastic motion generation with SVG and Monkey-Net using \textit{SSIM} and \textit{LPIPS} to measure visual realism and diversity respectively. }
\vspace{1mm}
\begin{tabular}{c|clcl}
\hline
\multirow{2}{*}{Method} & \multicolumn{4}{c}{\textit{Robot pushing}}                        \\ \cline{2-5} 
                        & \multicolumn{2}{c|}{SSIM ($\uparrow$)}    & \multicolumn{2}{c}{LPIPS ($\uparrow$)}  \\ \hline \hline
SVG                     & \multicolumn{2}{c|}{$0.815 \pm 0.006$} & \multicolumn{2}{c}{$0.0398 \pm 0.0005$} \\
Monkey-Net              & \multicolumn{2}{c|}{$0.783 \pm 0.008$}      & \multicolumn{2}{c}{N/A}      \\
Ours      & \multicolumn{2}{c|}{$\textbf{0.827}\pm 0.007$}      & \multicolumn{2}{c}{$\textbf{0.0422} \pm 0.0003$}      \\ \hline
\end{tabular}%
\vspace{-4mm}
\label{tab:i2v}
\end{table}

\subsection{Stochastic Motion Generation}\label{ssec:exp_sto}
\subsubsection{Qualitative results and comparisons.} In order to provide motion stochasticity in the produced video outputs, our Dual-MTGAN is able to generate frame-based motion features $z_{m,s}^t$ by sampling the video motion representation $z_M$ from a prior Gaussian distribution, and then synthesizes a corresponding output video based on a single input image. Example results are shown in Figure~\ref{fig:exp_i2v} on \textit{robot pushing} (left) and \textit{human actions} (right) datasets respectively, in which diverse outputs can be produced based on different sampled random vectors. From the above results, we see that our model is capable of synthesizing a set of plausible videos with reasonable motion by sampling different $z_M$ from $\mathcal{N}(\textbf{0},\textbf{I})$. 

To verify the plausibility of our stochastic motion generation model, we take the first frame of a video as the single input image, and compute \textit{Structural Similarity} (SSIM) between the generated videos and the ground truth video. We show the output videos with the best SSIM in the bottom rows of Figure~\ref{fig:exp_i2v}, and observe both visual and motion similarity between these outputs and their associated ground truth ones. Therefore, we confirm that our model is able to produce diverse yet realistic output videos.   

As shown in the \emph{right} part of Figure~\ref{fig:com}, we perform qualitative comparisons on \textit{robot pushing} dataset using SVG~\cite{denton2018stochastic}, Monkey-Net~\cite{siarohin2018animating}, and our model. Our model achieved improved temporal continuity and video quality across frames than \cite{denton2018stochastic} did and presented comparable visual fidelity as \cite{siarohin2018animating}. It is worth noting that SVG was not able to produce satisfactory results under the condition that only an input image was provided. 
In addition, Monkey-Net requires extra information (i.e., keypoints) for video generation, while ours is capable of synthesizing videos without the need to utilize pre-defined motion information. The quantitative comparison is available in supplementary material.

\begin{figure*}[t!]
	\centering
	\includegraphics[width=0.9\textwidth]{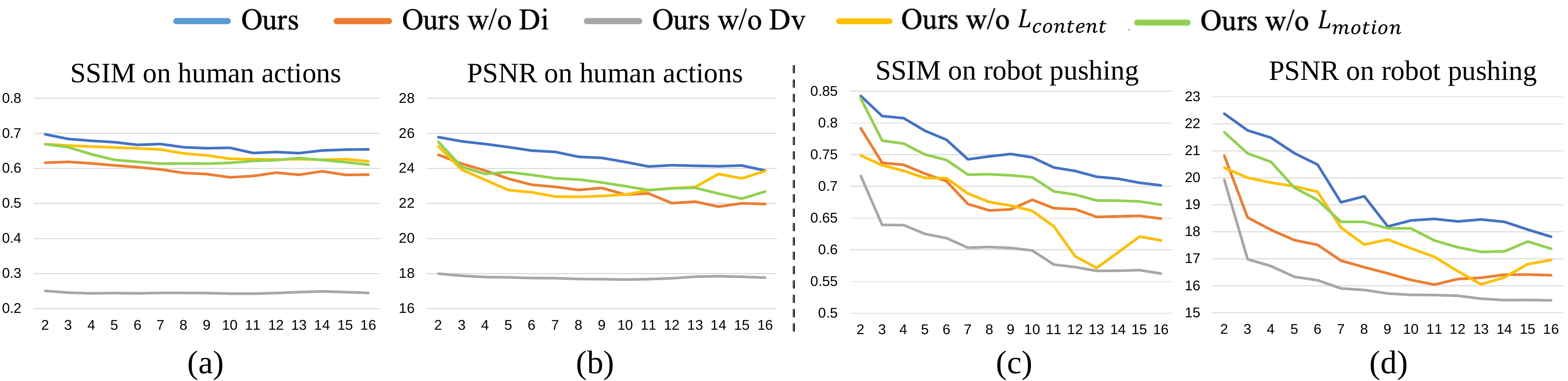}
	\vspace{2mm}
     \caption{Ablation studies of stochastic motion generation. Note that x and y axes indicating time step and SSIM (or PSNR) scores. (a) and (b) are evaluated on \textit{human actions}, while (c) and (d) present results on \textit{robot pushing} dataset. }
	\vspace{-4mm}
	\label{fig:quantitative}
\end{figure*}

\subsection{Quantitative comparisons}
We now conduct quantitative comparisons on \textit{robot pushing} dataset to measure both the visual realism and diversity using SSIM and LPIPS~\cite{zhang2018unreasonable}, respectively. We compute the best SSIM among all samples to measure the video quality, followed by~\cite{p2pvg2019}, and evaluate the diversity of generated videos by calculating average LPIPS distance between videos generated from the same target image, where higher LPIPS distance indicates outputs with more diversity. In Table~\ref{tab:i2v}, we
show that our model achieved satisfactory SSIM score comparing with SVG and Monkey-Net, and also surpassed SVG in terms of LPIPS distance. Note that, despite Monkey-Net is able to produce videos with fidelity, it only performs \textit{deterministic} video generation. Thus, the LPIPS distance for Monkey-Net is not considered in Table~\ref{tab:i2v}. With the above comparisons, we verify that our model is capable of synthesizing results with both plausibility and stochasticity.

\subsection{Ablation Study}\label{ssec:exp_abla}
We now provide ablation studies (more available in supplementary). For stochastic motion generation, we consider the \textit{human actions} and \textit{robot pushing} datasets, using our model 1) without $D_I$, 2) without $D_V$, 3) without $\mathcal{L}_{C}$, and 4) without $\mathcal{L}_{M}$. For each model, we produce 25 output videos, then compute the best \textit{Structural Similarity} (SSIM) and \textit{Peak Signal-to-Noise Ratio} (PSNR) scores with respect to the ground truth. As shown in Figure~\ref{fig:quantitative}, our model surpassed others in terms of both metrics, which confirms the video quality achieved by Dual-MTGAN. Note that, without the presence of the $D_I$, both scores dropped since there would be no guarantee for preserving image quality and its property. Moreover, when $D_V$ was disabled, both scores dropped drastically. This verifies that our $D_V$ has the ability to ensure that the produced videos exhibit both temporal continuity and visual quality.

Next, we further disable our temporal consistency $\mathcal{L}_{C}$ and motion consistency $\mathcal{L}_{M}$ to verify the effectiveness of our design for representation disentanglement. If the temporal consistency is disabled, we observe a huge drop on both scores. Without $\mathcal{L}_{C}$, the extracted content features are not trained to be time-invariant, and the image generator $G_I$ would synthesize frames with the entangled representations and ignore the motion features. Hence, we cannot derive a rich motion space to produce videos with sufficiently realistic dynamics and stochasticity. Finally, if the motion consistency term is not applied, the motion information would be omitted during the generation process, and thus lead to worse visual quality and lower SSIM and PSNR scores. 
With the above experiments, we confirm the effectiveness and robustness of our proposed Dual-MTGAN in performing image-to-video synthesis.
% As mentioned in Section~\ref{ssec:trans}, we apply adversarial learning in both image and video level to ensure the outputs would be generated with property and quality. To verify the effect of these modules, we do ablation study on the proposed framework and show the quantitative comparison in Figure~\ref{}. We adopt the metric, including PSNR and SSIM to measure the quality of the generated frames in the task of image-to-video generation. We claim that without image-based discriminator $D_I$, the image quality 

% As shown in Figure~\ref{}, if our model without video-based discriminator $D_V$, the SSIM and PSNR curve 

\section{Conclusion}\label{sec:conclusion}

In this paper, we proposed a unified deep learning model of Dual Motion Transfer GAN (Dual-MTGAN). This unique network addresses image-to-video synthesis, which is able to perform deterministic motion transfer or stochastic motion generation given an input image. This is realized by the design of encoders which disentangle temporal-coherent content and motion features, while the latter is modeled by a generative recurrent network modules. In our experiments, we successfully verified that our model performed promising deterministic motion transfer and stochastic motion generation results using facial expression, human actions, and robot pushing datasets with satisfactory visual quality and motion stochasticity.

\section*{Acknowledgment}
This work is supported in part by the Ministry of Science and Technology of Taiwan under grant MOST 109-2634-F-002-037.


%\section*{Acknowledgment}


\bibliographystyle{IEEEtran}
\bibliography{egbib}

\end{document}